\documentclass[review]{elsarticle}

\usepackage{lineno,hyperref}
\usepackage{amsmath}
\usepackage{amsfonts}
\usepackage{graphicx}
\usepackage{booktabs}
\usepackage{geometry}
\journal{arXiv}
\begin{document}

\begin{frontmatter}

\title{Transfer Learning Toolkit: Primers and Benchmarks}
\tnotetext[mytitlenote]{The toolkit is available at \url{https://github.com/FuzhenZhuang/Transfer-Learning-Toolkit}.}

%% Group authors per affiliation:
\author{Fuzhen Zhuang$^{1,2}$}
\author{Keyu Duan$^{1,2}$}
\author{Tongjia Guo$^{1,2}$}
\author{Yongchun Zhu$^{1,2}$}
\author{Dongbo Xi$^{1,2}$}
\author{Zhiyuan Qi$^{1,2}$}
\author{Qing He$^{1,2}$}
\address{$^1$Key Laboratory of Intelligent Information
	Processing of Chinese Academy of Sciences (CAS), Institute of Computing
	Technology, CAS, Beijing 100190, China.}
\address{$^2$University of Chinese Academy of Sciences, Beijing 100049, China.}
%\fntext[myfootnote]{Since 1880.}

%% or include affiliations in footnotes:
%\author[mymainaddress,mysecondaryaddress]{Elsevier Inc}
%\ead[url]{www.elsevier.com}

\begin{abstract}
The transfer learning toolkit wraps the codes of 17 transfer learning models and provides integrated interfaces, allowing users to use those models by calling a simple function. It is easy for primary researchers to use this toolkit and to choose proper models for real-world applications. The toolkit is written in Python and distributed under MIT open source license. In this paper, the current state of this toolkit is described and the necessary environment setting and usage are introduced.
\end{abstract}


\end{frontmatter}

%\linenumbers

\section{Introduction}
Transfer learning is a promising and important direction in machine learning, which attempts to leverage the knowledge contained in a source domain to improve the learning performance or minimize the number of labeled samples required in a target domain. According to the survey by Pan and Yang \cite{PY2010TKDE}, transfer learning approaches can be divided into four categories, i.e., instance-based, feature-based, parameter-based, and relational-based approaches. Instance-based approaches focus on re-weighting the instances in the source domain to help construct a learner on the target domain. Feature-based approaches aim to find a new feature representation for domain adaptation. Parameter-based approaches try to discover shared parameters or priors between the domains. Relational-based approaches build the mapping of knowledge across relational domains. As the development of deep-learning techniques, a number of transfer learning models have been constructed based on deep networks \cite{TSK2018ICANN}, which have shown excellent performance on a variety of tasks. 

In order to help primary researchers properly select and use some representative models as baselines in their comparative experiments with ease, the toolkit is developed, which contains a number of representative transfer learning models (wrapped or implemented by ourselves with the help of the existing open source code). The current version of this toolkit wraps 17 models and provides unified interfaces. These 17 transfer learning models include: HIDC \cite{ZLY2013IJCAI}, TriTL \cite{ZLD2014TC}, CD-PLSA \cite{ZLS2010CIKM}, MTrick \cite{ZLX2011SADM}, SFA \cite{PNS2010WWW}, mSDA \cite{CXW2012ICML}, SDA \cite{GBB2011ICML}, GFK \cite{G2012CVPR}, SCL \cite{BMP2006EMNLP}, TCA \cite{PTK2011TNN}, JDA \cite{LWD2013ICCV}, TrAdaBoost \cite{DYX2007ICML}, LWE \cite{GFJ2008KDD}, DAN \cite{LCW2015ICML}, DCORAL \cite{SS2016ECCVW}, MRAN \cite{ZZW2019NN}, and DANN \cite{GL2015ICML} \cite{GUA2016JMLR}. With this toolkit, users can simply call unified functions and run desired models, which may be helpful for exploring the transfer learning area or testing the superiority of new designed models. The models in the toolkit are temporarily divided into five groups, i.e., deep-learning-based, feature-based, concept-based, parameter-based, and instance-based groups. 

The rest of this paper is organized into four sections. Section \ref{sec:models} describes the models in the four parts, respectively. Section \ref{sec:es} introduces the environment settings. Section \ref{sec:ce} provides a case example. Section \ref{sec:cfw} presents the conclusion and the future work.

\section{Models}
\label{sec:models}
Several definitions about transfer learning are presented below \cite{PY2010TKDE,ZQD2019ARXIV}.

$\boldsymbol{(Domain)}$. {\it A domain $\mathcal{D}$ is composed of two parts, i.e., a feature space $\mathcal{X}$ and a marginal distribution $P(X)$. In other words, $\mathcal{D}=\{\mathcal{X},P(X)\}$. And the symbol $X$ denotes an instance set, which is defined as $X=\{\mathbf{x}|\mathbf{x}_i\in\mathcal{X},~i=1,\cdots,n\}$.}

$\boldsymbol{(Task)}$. {\it A task $\mathcal{T}$ consists of a label space $\mathcal{Y}$ and a decision function $f$, i.e., $\mathcal{T}=\{\mathcal{Y},f\}$. The decision function $f$ is an implicit one, which is expected to be learned from the sample data.}

Transfer learning utilizes the knowledge implied in the source domain to improve the performance of the learned decision function on the target domain \cite{PY2010TKDE,ZQD2019ARXIV}. For example, a common scenario of transfer learning is that we have abundant labeled instances in the source domain but only a few or even none of labeled instances in the target domain \cite{ZQD2019ARXIV}. In such condition, the target of a transfer learning task is to build a more efficient decision function on the target domain with the data from both the source and the target domains.

Given that some models require the labeled instances, while others do not. In order to unify the calling function of models, the interface is designed as follows, i.e.,
$$
{\text{accuracy}} = {\text{calling\_function}}(X_{\text{s}}, X_{\text{t}}, X_{\text{test}}, Y_{\text{s}}^*, Y_{\text{t}}^*, Y_{\text{test}}),
$$
where $X_{\text{s}}$ and $ X_{\text t}$ denote the instances in the source and the target domains, respectively; $X_{\text{test}}$ and $ Y_{\text{test}}$ denote the instances and the corresponding labels used for test; $Y_{\text s}^*$ and $Y_{\text t}^*$ refer to the labels of $X_{\text{s}}$ and $ X_{\text t}$, respectively. Note that $^*$ means that the two inputs are optional, which depends on the corresponding model. 
For example, the calling function in TCA \cite{PTK2011TNN} is 
$$
{\text{accuracy}} = {\text{TCA.fit\_predict}}(X_{\text{s}}, X_{\text{t}}, X_{\text{test}}, Y_{\text{s}}, Y_{\text{test}}),
$$
while the interface in TrAdaBoost \cite{DYX2007ICML}, which requires the labels of the instances in target domain, is given by
$$
{\text{accuracy}} = {\text{TrAdaBoost.fit\_predict}}(X_{\text{s}}, X_{\text{t}}, X_{\text{test}}, Y_{\text{s}},Y_{\text{t}}, Y_{\text{test}}).
$$
The models in the toolkit are also divided into two general groups, i.e., deep-learning-based and traditional (non-deep-learning) groups. The traditional models are further categorized into feature-based, concept-based, parameter-based, and instance-based ones (corresponding to the four file folders in the toolkit). Besides, the traditional models are wrapped into classes, and the hyper-parameters of these models can be customized by setting the initial function. If their parameters are not altered, the parameters will be set to default. Table \ref{tab:categorization} shows the categories of the models contained in the toolkit.
\begin{table}[!t]
	\caption{Categorization used in the toolkit}
	\centering
	\label{tab:categorization}
	\begin{tabular}{ll}
		\toprule
		Category     & Models  \\
		\midrule
		Feature-based & SFA \cite{PNS2010WWW}, mSDA \cite{CXW2012ICML}, SDA \cite{GBB2011ICML}, GFK \cite{G2012CVPR}, SCL \cite{BMP2006EMNLP}, TCA \cite{PTK2011TNN}, JDA \cite{LWD2013ICCV}\\
		Concept-based     & HIDC \cite{ZLY2013IJCAI}, TriTL \cite{ZLD2014TC}, CD-PLSA \cite{ZLS2010CIKM}, MTrick \cite{ZLX2011SADM} \\
		Parameter-based    & LWE \cite{GFJ2008KDD} \\
		Instance-based    & TrAdaBoost \cite{DYX2007ICML} \\
		Deep-learning-based & DAN \cite{LCW2015ICML}, DCORAL \cite{SS2016ECCVW}, MRAN \cite{ZZW2019NN}, DANN \cite{GL2015ICML}\cite{GUA2016JMLR}\\
		\bottomrule
	\end{tabular}
\end{table}

Feature-based models in the toolkit contain SFA \cite{PNS2010WWW}, mSDA \cite{CXW2012ICML}, SDA \cite{GBB2011ICML}, GFK \cite{G2012CVPR}, SCL \cite{BMP2006EMNLP}, TCA \cite{PTK2011TNN}, and JDA \cite{LWD2013ICCV}. These models, which are the representative ones in traditional transfer learning, mainly focus on altering the feature representations, i.e., transforming the instances in the original feature space to a designed new feature space. In terms of these models, several additional interfaces are designed including ${\text{fit()}}$, ${\text{transform()}}$, etc. It is worth mentioning that some of these models, i.e., TCA \cite{PTK2011TNN}, JDA \cite{LWD2013ICCV}, and GFK \cite{G2012CVPR} are implemented and wrapped based on the repository constructed by Wang {\it et al} \cite{W0000XYZ}. Besides, SDA \cite{GBB2011ICML} and mSDA \cite{CXW2012ICML} are implemented and wrapped based on \cite{MAD0000SDAE} and \cite{DOU0000MSDA}.  

Concept-based models in the toolkit include HIDC \cite{ZLY2013IJCAI}, TriTL \cite{ZLD2014TC}, CD-PLSA \cite{ZLS2010CIKM}, and MTrick \cite{ZLX2011SADM}. The source code of these four models is written in MATLAB and has been wrapped by Dr. Cheng into a MATLAB toolbox, i.e., TLLibrary64 \cite{ZCL2015IJCAI}. The interfaces in the toolkit call the entry function of that toolbox directly. Note that the current MATLAB engine API for Python only supports Python 2.7, 3.5, and 3.6. If the Python version is higher than 3.6, the Python interfaces may not be executable. The MATLAB source code is included in the toolkit so that the user may directly call the functions in MATLAB.

The parameter-based and the instance-based models in the toolkit contain LWE \cite{GFJ2008KDD} and TrAdaBoost \cite{DYX2007ICML}, respectively. The model of TrAdaBoost \cite{DYX2007ICML} is implemented and wrapped based on \cite{CHEN0000TAB}. Note that the experimental results of LWE model in the toolkit are not consistent with the results in the original paper \cite{GFJ2008KDD}, because the clustering methods used are different. To reproduce the results in paper \cite{GFJ2008KDD}, please use CLUTO \cite{K0000CLUTO}, which is a data clustering software tool written in C.

The deep-learning-based models in the toolkit include DAN \cite{LCW2015ICML}, DCORAL \cite{SS2016ECCVW}, MRAN \cite{ZZW2019NN}, and DANN \cite{GL2015ICML} \cite{GUA2016JMLR}. The code is from \cite{ZW0000GIT}, which is a repository containing a number of deep transfer learning implementations. Each deep-learning-based model in the toolkit is implemented by three files, i.e., model.py, loss.py, and main.py. To use a new dataset, a data loading file is necessary. The sample files are provided in the toolkit. 

\section{Environment Setting}
\label{sec:es}
In this section, the environment setting steps are introduced. To use the toolkit, a Python environment is necessary and the version 3.6 is recommended because the Python code in the toolkit is written by Python 3. If the Python version is 3.7 or higher, the Python interfaces for concept-based models may not be executable because the official MATLAB engine API for Python only supports version 2.7, 3.5, and 3.6. The support is valid for versions prior to R2019b. For the versions of Python not supported, the MATLAB entry functions of concept-based models are provided. The MATLAB environment should be prepared in advance in order to use the concept-based models. There are two steps. First, add the directory of utilities to the MATLAB paths. Then, install MATLAB engine API for Python by using the following commands.
\begin{align}
& {\text{cd \ ``matlabroot/extern/engines/python''}} \notag\\
& {\text{python \ setup.py \ install}} \notag
\end{align}

\section{Case Example}
\label{sec:ce}
To introduce the usage of the toolkit, TCA \cite{PTK2011TNN} is used to show the calling process. After the preparation of the running environment and the datasets, the corresponding class should be imported from TCA.py and the parameters should be initialized. Then, the entry interface is called to run the model. The code is as follows.
\begin{align}
& {\text{import \ TCA \ from \ TCA}} \notag\\
& {\text{model = TCA(parameters = settings)}} \notag \\
& {\text{accuracy = model.fit\_predict}}(X_{\text{s}}, X_{\text t}, X_{\text{test}}, Y_{\text s}, Y_{\text{test}}) \notag
\end{align} 

Reuters-21578\footnote{\url{https://archive.ics.uci.edu/ml/datasets/Reuters-21578+Text+Categorization+Collection}}, which is a hierarchical dataset for text categorization, is used to test the performance of TCA \cite{PTK2011TNN}. The results are given in Table \ref{tab:results}. Besides, the experimental results of all the models are available in \cite{ZQD2019ARXIV}.
\begin{table}[!t]
	\centering
	\caption{Experimental results of TCA.}
	\setlength{\tabcolsep}{1.8mm}
	\begin{tabular}{cccc}
		\toprule
		Model & Orgs vs Places & People vs Places & Orgs vs People \\ 
		\midrule
		TCA & 0.7368 & 0.6065 & 0.7562 \\
		\bottomrule
	\end{tabular} 
	\label{tab:results}
\end{table}

\section{Conclusion and Future Work}
\label{sec:cfw}
In this paper, a new toolkit has been introduced, which contains 17 representative transfer learning models. In the toolkit, several unified interfaces are provided, which makes it easy for primary researchers to use. The toolkit is written in Python 3 and makes use of a MATLAB toolbox. In the future, we will keep on maintaining, improving, and enriching the toolkit. New representative models will be added to the toolkit and various code languages will be unified. 

If you are interested in this toolkit and use it for baselines, please cite this paper and our survey paper \cite{ZQD2019ARXIV}. Also, feel free to contact us if you have any problem about using this toolkit. \\ (Fuzhen Zhuang: zhuangfuzhen@ict.ac.cn, Keyu Duan: Karen\_Duane@buaa.edu.cn)

\end{document}